%% file: 0-main.tex
\documentclass[sigconf]{acmart}

\copyrightyear{2022} 
\acmYear{2022} 
\setcopyright{acmcopyright}\acmConference[AIES'22]{Proceedings of the 2022 AAAI/ACM Conference on AI, Ethics, and Society}{August 1--3, 2022}{Oxford, United Kingdom}
\acmBooktitle{Proceedings of the 2022 AAAI/ACM Conference on AI, Ethics, and Society (AIES'22), August 1--3, 2022, Oxford, United Kingdom}
\acmPrice{15.00}
\acmDOI{10.1145/3514094.3534177}
\acmISBN{978-1-4503-9247-1/22/08}

\usepackage{multirow}
\usepackage{booktabs}
\usepackage{subfiles}
\usepackage{ifthen}
\usepackage{color}
\usepackage{wrapfig}
\usepackage{listings}
\usepackage{xcolor}
\usepackage{tikz}
\usepackage{listings}
\usepackage{amsmath}
\usepackage{natbib}
\usepackage{qtree}
\usepackage{tcolorbox}
\usepackage{dirtytalk}
\usepackage{multicol}
\usepackage[symbol]{footmisc}

\newcommand{\xhdr}[1]{\vspace{3mm}\noindent{{\bf #1.}}}

\newcommand{\xhdrsub}[1]{\noindent{\textit{#1.}}}

\newcommand{\showComments}{false}

\newboolean{comments}
\setboolean{comments}{\showComments}
\ifthenelse{\boolean{comments}} {
	\newcommand{\sid}[1]{\textcolor{blue}{(Sid: #1)}}
	\newcommand{\reid}[1]{\textcolor{magenta}{(Reid: #1)}}
	\newcommand{\ashton}[1]{\textcolor{red}{(Ashton: #1)}}
	\newcommand{\jon}[1]{\textcolor{green}{(Jon: #1)}}
	\newcommand{\solon}[1]{\textcolor{cyan}{(Russell: #1)}}
} {
	\newcommand{\sid}[1]{}
	\newcommand{\reid}[1]{}
	\newcommand{\ashton}[1]{}
	\newcommand{\jon}[1]{}
	\newcommand{\solon}[1]{}
}



\begin{document}
\fancyhead{}

\title{Mimetic Models: Ethical Implications of AI that Acts Like You}

\author[R. McIlroy-Young]{Reid McIlroy-Young}
\email{reidmcy@cs.toronto.edu}
\affiliation{%
  \institution{Department of Computer Science\\University of Toronto}
  \city{Toronto}
  \state{Ontario}
  \country{Canada}
}

\author[J. Kleinberg]{Jon Kleinberg}
\affiliation{%
  \institution{Department of Computer Science\\
  Cornell University}
  \city{Ithica}
  \state{New York}
  \country{USA}
}

\author[S. Sen]{Siddhartha Sen}
\affiliation{%
  \institution{Microsoft Research}
  \city{New York City}
  \state{New York}
  \country{USA}
}

\author[S. Barocas]{Solon Barocas}
\affiliation{%
  \institution{Microsoft Research\\ \& Cornell University}
  \city{New York City}
  \state{New York}
  \country{USA}
}

\author[A. Anderson]{Ashton Anderson}
\affiliation{%
  \institution{Department of Computer Science\\University of Toronto}
  \city{Toronto}
  \state{Ontario}
   \country{Canada}
}

\begin{abstract}
\subfile{1-abstract}
\end{abstract}

\maketitle

\subfile{2-intro}

\subfile{4-scenarios}
\subfile{5-ethical-discussion}

\subfile{6-related-works}

\subfile{7-conclusion}

\bibliographystyle{acm}
\bibliography{00-cites}

\end{document}

%% file: 1-abstract.tex
An emerging theme in artificial intelligence research is the creation of models to simulate the decisions and behavior of specific people, in domains including game-playing, text generation, and artistic expression. These models go beyond earlier approaches in the way they are tailored to individuals, and the way they are designed for interaction rather than simply the reproduction of fixed, pre-computed behaviors. We refer to these as {\em mimetic models}, and in this paper we develop a framework for characterizing the ethical and social issues raised by their growing availability. Our framework includes a number of distinct scenarios for the use of such models, and considers the impacts on a range of different participants, including the target being modeled, the operator who deploys the model, and the entities that interact with it.

%% file: 2-intro.tex
\newcommand{\nhdr}[1]{\subsection {#1}}
\newcommand{\yhdr}[1]{\vspace{1mm}\paragraph{\bf {#1}.}}

\newcommand{\omt}[1]{}

\section{Introduction}

When machine learning (ML) is deployed to replace human effort on tasks in specific application domains, the primary focus has traditionally been on the performance of the ML system relative to human capability on the relevant tasks~\cite{mnih2015human,silver2016mastering,silver2018general,moravvcik2017deepstack,jaderberg2019human}; but there has been increasing interest in trying to design ML solutions that exhibit human-like behavior on the task, generating solutions that look like what a skilled human being would produce~\cite{mcilroy2020aligning,jacob2021modeling,bard2020hanabi,liang2019implicit}. In domains where there is extensive data on individual behavior, it becomes possible to build such models not simply on aggregate human behavior, but tailored to the behavior of {\em specific (individual) people}---a model that tries to simulate the actions of a particular person in arbitrary situations within the domain. 

The idea of designing ML models to simulate specific people is becoming a reality in a growing number of domains---particularly for game-playing, where chess engines have been trained to play like specific human chess players~\cite{mcilroy2020learning}, and e-sports avatars have been trained to play like specific human athletes~\cite{Thomas2010Nov,Morris2013Aug,Tantaros2019Jan}; and for writing and text generation, where models have been trained to produce text in the writing style of specific authors~\cite{bird2020lstm} or social media users~\cite{ressmeyer2019deep}. From these realized examples, it becomes possible to see how the same techniques could be used in other forms of artistic expression (for example, to compose music in the style of specific people), or professional expertise (where work in medical AI is beginning to explore the design of models that try to match the diagnoses of specific doctors \cite{guan2018said}). The concreteness of these developments makes clear that it is an appropriate time to identify the common themes across the efforts in these different domains, and to consider their ethical and social implications. 

With this in mind, we define a {\em mimetic model} to be an algorithm that is trained on data from a {\em specific individual} in a given domain, and which is designed to accurately predict and {\em simulate} the behavior of this individual in {\em new situations} from the domain. This definition is intended to capture the examples discussed above, and to highlight the key themes that we believe are central to them. Crucially, a mimetic model is {\em generative} in the sense that it does not exist simply to predict a specific person's behavior, but to produce this behavior in context and thus interact with new environments and new individuals. In this way, the mimetic model is broader than any one of its outputs; it is not simply an e-mail message, tweet, or chess move that looks like it was created by you, but a mechanism that can be placed in arbitrary situations and produce messages, tweets, or chess moves that are designed to resemble what you would do in these situations. 

\subsection{Mimetic Models: Analyzing their Ethical and Social Implications}\label{sec:mm}

Within this framework, we ask: What normative issues come into play when mimetic models enter more widespread use across a diverse set of domains? In posing this question, we note that the normative impacts can be both produced and experienced by several different parties: the {\em target} individual that the mimetic model is designed to simulate; the {\em creator} who builds the model and the {\em operator} who uses it; and finally the party who {\em interacts} with the model. 

One of the most immediate concerns about a mimetic model is its potential to be used for deception: someone could believe that they are interacting with you when they are interacting with a mimetic model of you. These concerns are related to the role that deepfakes play in spreading misinformation~\cite{vaccari2020deepfakes}, and we discuss the relationships and distinctions with mimetic models further in Section \ref{sec:deepfake}. 

But an animating motivation of the present paper is to realize how many additional normative concerns remain even when a mimetic model is not being used deceptively---that is, even when there is transparency about which agents in a given domain are mimetic models, and which target individuals they are based on. We organize these concerns into three broad categories, which we analyze in the subsequent sections of the paper. 
To give a sense for some of the questions that motivate this analysis, we begin with an overview of the three categories and some of the issues that arise in each.

\yhdr{Mimetic Models as a Means} First, interacting with a mimetic model can be used as preparation for interactions in real life---essentially, as a means to an end, where the end is the real-life interaction. For example, in a setting where two people $A$ and $B$ are going to meet so that $B$ can interview $A$ for a job, we could imagine that $A$ might ``practice'' the interview dozens of times with a mimetic model of $B$ so as to find the types of answers that seem most appealing to $B$. And in place of a job interview, we could imagine that $A$ is repeatedly practicing their interactions with $B$ in preparation for a journalistic interview, for a fund-raising pitch, or for going on a date. You might feel differently, for example, being interviewed by a journalist if you knew they had spent several days practicing their interview on a mimetic model of you. And in a competition such as a chess game, where we've noted that mimetic models are already feasible, $A$ could practice against a mimetic model of a future opponent $B$ to identify their weaknesses. 

We might reasonably feel that there are qualitative contrasts between the way this type of preparation operates across different domains---for example, rehearsing a chess game against a mimetic model may raise different concerns than rehearsing a conversation with a future date. But in all cases, the availability of mimetic models would change the underlying norms and expectations that we would have for one-on-one interactions between people, the expectations for what people should disclose about the context of these interactions, and the conclusions we draw from them as a result.

\yhdr{Mimetic Models as an End} Another broad category of uses we envision are those in which a mimetic model is used not to prepare for future interactions, but as an end in itself. In some cases, this end might be the potential for interaction with the mimetic model. For example, chatbots have been used as automated teaching assistants in online class forums \cite{chopra2016meet}, but these have tended to be automated agents trained on aggregate data. We could ask how the normative considerations might change if an online class created a mimetic model of each human TA, so that students who preferred the style of office-hour help offered by human TA $A$ could choose their mimetic model over the model of human TA $B$. This is a form of work replacement through automation that is highly personalized, and raises questions about how human labor can become devalued, about the forms of consent and compensation the real $A$ and $B$ are entitled to, and about the responsibilities the model creators have to faithfully represent the behaviors of $A$ and $B$, given that will be presented to the world through their mimetic models. If the mimetic model of $A$ is a highly accurate representation but also rude to students, should that reflect poorly on the real $A$? If students develop a social dynamic in which they are verbally abusive on the forum to the mimetic model of $A$, what harms has the real $A$ suffered as a result? 

Mimetic models can also be used as an end in themselves when they are deployed to satisfy the interest of a group that functions as an audience or a set of spectators. For example, ``what if'' scenarios are a source of fascination among fans of all sports---in a given sport, what if $A$, the greatest player of the 1980s, had had the opportunity to play at their peak against $B$, the greatest player of the present day? Mimetic models provide a new mechanism for such thought experiments, and they raise analogous questions of reputation, compensation, and consent. If a mimetic model of former chess champion $A$ defeats a mimetic model of former chess champion $B$, what do we expect to be the reputational consequences in practice for the real $A$ and $B$? And what stake should $A$ and $B$ have in the creation and use of their mimetic models for these purposes?  More narrowly scoped precursor questions have been the subject of litigation in the video-game domain, when statistics and likenesses of athletes have been used without consent or compensation \cite{cimini2018walking}; the increasing fidelity of mimetic models has the potential to intensify all of these considerations. 

\yhdr{Mimetic Models of Oneself} Many of the normative considerations involving the use of mimetic models depend on the relationship of the model's target to its creator and/or operator. The cases above make clear how diverse this set of relationships can be. 
But a relationship that brings up a specific set of considerations, and therefore benefits from separate analysis, is the case in which the operator of the mimetic model is its target---that is, the case in which someone builds a mimetic model of themself.

There are several natural uses for a mimetic model of oneself. One of the most basic is as a {\em force multiplier}; for example, a model trained to generate e-mail replies in your style could be used to answer significantly more messages than you are practically able to handle on your own. This raises questions about the level of disclosure that is appropriate for the authorship of such messages, and how norms about appropriateness will evolve. A world in which e-mail messages written by a mimetic model are explicitly flagged as such may produce different social cues than a world in which it is left ambiguous which messages were authored by the real you and which by your mimetic model. The use of effort as a means of signaling commitment to a relationship would look different in these two worlds. And even under a norm where people explicitly flag messages that were written by their model, there are more subtle choices about what to disclose and what to reveal. Is it deceptive, for example, for you to use a mimetic model whose level of politeness has been covertly increased to a level beyond your own natural politeness? Is this fundamentally more deceptive than manually following advice, without the use of a model, for how to write e-mail that sounds more polite? 

\nhdr{Framework and Related Concepts}

With this range of potential scenarios in mind, it is useful to return to the general properties that characterize a mimetic model. As discussed above, there are three crucial aspects to this type of model: (i) it is targeted to a specific individual, rather than attempting to simulate human behavior in an aggregate sense; (ii) it is generative, in that it produces new behaviors; and (iii) it is interactive, in that it generates these behaviors in response to interactions with other individuals or with new environments. 

This structure, as well as the scenarios above, make clear that there are four roles that are important in any mimetic model: 
\begin{itemize}
\item The {\em target}, whose behavior the mimetic model is designed to simulate. We will say that the {\em fidelity} of the mimetic model is its accuracy in matching the behavior of the target. 
\item The {\em creator}, who builds the mimetic model. This implies that, at some level, the creator has at least indirect access to data about the target. 
\item The {\em operator}, who uses the mimetic model. 
\item The {\em interactor}, who engages in some form of interaction with the mimetic model. In different scenarios, the interactor might be communicating with the mimetic model, competing against it, or potentially watching it as a spectator. 
\end{itemize} 

Because AI systems interact with human behavior in so many different ways, it is also useful to situate the notion of mimetic models in comparison to related concepts. Of course, different concepts will naturally blend into each other, and so some of the distinctions that we draw here are questions of degree rather than absolute contrasts. 

First, {\em recommendation systems} naturally depend on personalized models of their users~\cite{bobadilla2013recommender,resnick1997recommender}. We think of these as distinct from the general formulation of mimetic models in that user models for recommendations tend to be focused on the narrow task of predicting a user's preferences for particular pieces of content, and providing content that is likely to satisfy the user. 

In a different direction, {\em deepfakes} are a type of manipulated media---often video---designed to portray specific people engaging in behaviors that didn't occur in real life~\cite{deepfake}. These can be used in deceptive or defamatory ways, or in instances where for example a deepfake of an actor's image or voice is used in a movie where they could not appear~\cite{deepfake_report}. Deepfakes clearly raise a number of ethical considerations that parallel what we consider for mimetic models, but it is important to note the key distinction that deepfakes tend not to be designed for unrestricted interaction with their environment or with others, but rather to present a static, precomputed set of behaviors. 

We discuss these comparisons, as well as additional related concepts, in Section \ref{sec:related} later in the paper. We turn next to a more in-depth discussion of our main categories of scenarios. Throughout our analysis, we focus on characterizing the novel ethical and social questions that mimetic models raise. Fully addressing these questions will likely involve a large collective effort over a number of years.

%%%%%%%%%%%%%
%%% Previous version, now commented out
%%%%%%%%%%%%%

\omt{
\section{Introduction}

%I can add cites if we like the examples --Reid
A recent trend in machine learning papers is the creation of models that \textit{act} like specific people. This takes a wide variety of forms, from researches attempting to reconstruct potential historical events by modeling the participants, game player's styles being captured, or extensions to auto-complete that attempt to respond to emails like their user.

These models raise unique ethical questions, e.g. "Is it ethical to attribute theoretical actions to people, based on historical data?", "How should the researchers handle models of specific game players?", and "What are the potential negative outcomes of creating an 'better' version of yourself to do a task?". These and many more that we will outline here that are unique to theses "mimetic models", will be discussed here. There are of course many other ethical questions raised by these models which are shared by other types of modeling: such as those regarding deception, and prediction, which have been well discussed in other literature. We are here focused on the novel ethical issues.

To better understand and illustrate the ethical issues, we will present a series of scenarios. These are all based on contemporary research, we are not attempting to be Cassandra. We can consider how a mimetic model can be used by a single person to model themselves, e.g. an auto-complete system. The goal of these models is to scale yourself out, but if the model is an imperfect replica that could lead to it creating a bad impression of you, or possibly worse it could outperform you on its narrow task setting an impossible standard for you. An other interesting potential use of mimetic models is for someone to understand another person's approach to some task. This could be a game of chess, where you want to learn your opponent's style, or preparation for a date, rehearsing so that the night will go perfectly--as you intend. The computational aspect of the mimetic models means that they could be distributed, consider a tournament where people pay to play against models of their favorite sports stars. How does that effect the value of the start's labour, and what if the models are done without their permission. This issue has already had multiple court cases, so this is not ideal speculation. There is also the use of mimetic models as teaching aids, what is the ethical obligation of a teacher to their students? Can they make a model of each of them to predict the best lesson? What if the model is not perfect. Finally, we can consider the mimetic models being used to populate a world, where the operator simply wants diversity in some virtual world. This has been a goal of videogame AI designers for decades, making a world where every virtual character has a unique personalty. Do they own anything to the people they are using as fodder?

\reid{TODO add closing}

\subsection{Old Intro}
The authors of the recent paper~\cite{mcilroy2020aligning}, which presents a method for creating models that predict the next decision (move) specific players will make, realized when attempting to release the code and models that they could have some negative impacts on the players they had be testing on. This discussion\footnote{which led to reaching out to us} lead to the realization that the type of models they created raise ethical issues that the current literature does not address. These issues are being encountered right now and will become more common as techniques improve.

The main question of the McIlory-Young, et al paper is: "How should the researchers handle models of specific chess players?" these models are based on public data so the players are likely unaware of the models, but if released the models are trivially easy to run and then could be used to identify the source players.

This discussion inspired us to work towards characterizing what is novel about these concerns. To that end we examine the current state of the art and see that McIlory-Young, et al~\cite{mcilroy2020learning} are not alone. There are other works that have similar issues~\cite{Zhange2e,Welch2018May,guzdial2019friend}, but so far they have not been connected. We also see that work in the space is ongoing, with as time progresses more and more of these models will be created. So having an ethical framework now before major harm can happen is important. To that end we introduce a new term to describe these systems: "\textit{Mimetic Models}", describing how they are situated relative to other related concepts like personalization~\cite{mcauley2022personalized}, deepfakes~\cite{deepfake}, brain uploading\footnote{A concept from science fiction}~\cite{brin_2002,egan_2002,vinge_monster}, etc. Then we discuss the state of the art, where mimetic models are being used currently, and where work is ongoing. Next we discuss different contexts in which they can be used, both as they currently exist and how we expect them to present in the near future. These then act as a taxonomy in which to categorize and examine the different possible harms these models may pose. Clear discussion is required as many parties to the model may be harmed, be they the target of the model who's value is diminished, the counter-party interacting with it who is mislead, or even the operator running the system who's results are unwittingly biased.

}

%% file: 4-scenarios.tex
\section{Applications of Mimetic Models}

One way to classify the potential applications of mimetic models is to start by considering the possible ways through which an event in the world might be affected by the existence of such a model. 
At a high level, the event might be affected because an individual arrives at the event better prepared through their prior interaction with a mimetic model;
or the event might be affected because the mimetic model directly participates in the event.

In the former case, we think of the mimetic model as a means to an end, in that it prepares someone for a future interaction but is not necessarily present when the event takes place.
For example, a mimetic model might be used to help people prepare to interact with the actual person who is the target of the model. In this case, mimetic models would serve as a way for people to learn how best to achieve their goals in interacting with a person by first interacting with the mimetic model of the person.

In the latter case, when the mimetic model directly participates in the event, such a model could potentially be used as a complete substitute for the person who is the target of the model.  
For such cases, we make a further distinction between (i) scenarios in which an event that could have occurred with the genuine target instead takes place with the model, and (ii) counterfactual scenarios that could not feasibly have occurred without the presence of a mimetic model: for example, scenarios in which mimetic models of athletes or artists from different eras interact with one another---a type of interaction that could not have happened in real life.

Across all of these scenarios, we also consider the special case where the target and operator of a mimetic model is the same person. In such cases, a person might use a model of themself as a means to an end, having the model explore the world on the person's behalf to help the person better prepare to act in it themself; and as an end in itself, offloading certain tasks that the person would have otherwise needed to perform themself.

% Earlier version:
% At a high level, we imagine mimetic models being used for two quite different purposes. First, they might be used to help people prepare to interact with the actual person who is the target of the model. In this case, mimetic models would serve as a means to an end, a way for people to learn how best to achieve their goals in interacting with a person by first interacting with the mimetic model of the person. Second, they might be used as a complete substitute for interacting with the person who is the target of the model, not as a way to prepare for future interactions. In this case, allowing people to interact with the mimetic model in place of the actual person would be a valuable end in itself. We then consider the special case where the target and operator of a mimetic model is the same person. In such cases, a person might use a model of themself as a means to an end, having the model explore the world on the person's behalf to help the person better prepare to act in it themself; and as an end in itself, offloading certain task that the person would have otherwise needed to perform themself.

To give each of these possibilities greater substance, we consider a range of more concrete scenarios that illustrate how this might work, with some scenarios already visible in practice, others practically feasible, and still others being possibly feasible in the (perhaps distant) future. In progressing through these different scenarios, we hope to highlight the different ethical issues that different uses of mimetic models might raise.  We organize the section based on the distinctions discussed above, beginning with mimetic models as a means to an end (Section \ref{sec:means}), then as an end in themselves (Section \ref{sec:ends}), and finally for the case in which an individual creates a model of themself (Section \ref{sec:self}).

\subsection{Mimetic Modelling as a Means to an End}\label{sec:means}
We first consider the ways in which mimetic models might be used as a means to an end---that is, as a way to learn about the target of the mimetic model so as to be better able to achieve certain goals when interacting with the actual person in the future.

%In particular, we'll consider a range of scenarios in which mimetic models make it possible to anticipate how the target would respond to different actions and the ethical issues that such capabilities might raise. 

%The most directly ethically challenging use of \textit{mimetic models} is to understand the behaviour of other people.

\xhdr{Preparing for a competition} Imagine a person who has access to a mimetic model of a future opponent that they hope to defeat in an upcoming chess tournament. Further imagine that the person can rely on the mimetic model of their opponent to see how the opponent would respond to different moves and strategies. For example, to prepare to play the opponent at the tournament, the person could play as many games against the mimetic model as time allows. The person could also see how the mimetic model would respond to specific positions, rather than playing a full game linearly to its conclusion. Or the person could make a move, see how the mimetic model responds, and, if the move did not have the anticipated benefit, take back the move to try an alternative to see if that would be any more successful. The person could even have super-human (i.e., non-mimetic) chess-playing agents play against the mimetic model of their opponent to discover weaknesses that the person would not have even thought to test for.

Such a scenario is not fantastical. Recent research has demonstrated that it is possible to build mimetic models of particular players when there are available records of people's past game play~\cite{mcilroy2020aligning,mcilroy2020learning,Thomas2010Nov,Morris2013Aug,Tantaros2019Jan,gitlin2021Dec,dhou2018towards,tomavsev2020assessing,mnih2015human, silver2018general,bard2020hanabi,jacob2021modeling}, whether we're considering Chess~\cite{tomavsev2020assessing}, Go~\cite{mnih2015human}, Shogi~\cite{silver2018general}, Hanabi~\cite{bard2020hanabi}, Diplomacy~\cite{jacob2021modeling}, or other games with a finite set of legal moves. In these games, player actions can be recorded with perfect accuracy. Relying on players' past games as training data, it is thus possible to create a deep learning-based model that would likely make the moves of specific players. Recent work shows that building such player-specific models is even possible with a rather small sample of a player's past games~\cite{mcilroy2020learning}.

%\schdr{Mimetic Game Playing} Consider a game where players select actions from a finite pool of legal moves, e.g. 

%This is already being done, although in limited scopes such as game playing~\cite{mcilroy2020aligning,mcilroy2020learning,Thomas2010Nov,Morris2013Aug,Tantaros2019Jan,gitlin2021Dec}, or as an option in large language models~\cite{char-rnn,Radford2021Feb, brown2020language}.

What ethical issues does such a scenario raise? In particular, what, if anything, is different about a person preparing to play an opponent by looking over the opponent's publicly available past game play, which is common practice in competitive chess, and playing a mimetic model of the opponent? What advantage, if any, does the mimetic model give the person preparing for this match in comparison to the more traditional ways that a person might prepare?\footnote{Participants having too much information about competitors' strategies (solving the `metagame'~\cite{morgenstern1953theory,kokkinakis2021metagaming}) in a tournament is something that tournament operators already know to guard against~\cite{mtgMetaSolved}, since it degrades the experience for participants and observers by reducing the diversity of strategies.} %How do mimetic models change this?
One way to try to answer this question is to compare how the person learns under these two different scenarios. When a person is trying to learn from an opponent's past game play, they must expend considerable effort reviewing all of their opponent's past game play and attempt to generalize from these examples---that is, to not only memorize how the opponent has acted in the face of specific positions, but to induce a rule from past game play that would indicate how the opponent would act in the face of previously unencountered positions. Reliably extrapolating from an opponent's past game play is a non-trivial task both in terms of the time that must be invested by the person and the cognitive demands placed on them. A mimetic model would essentially do this work for the person: it would generalize from the opponent's past game play, relieving the person of the burden of manually studying individual games and positions, while also likely exceeding the person's ability to generalize accurately from these past examples. Note, however, that while the mimetic model might have assumed these burdens, the person still needs a way to learn from the lessons that the mimetic model has drawn from the opponent's past game play. The obvious way that the person might try to do this is to play games against the mimetic model or see how the mimetic model responds to specific positions, as described above. This then raises the question of whether learning about an opponent by playing a mimetic model of them is a more effective or efficient way to prepare for playing them than simply reviewing the opponent's past game play. As mentioned, there is good reason to believe the mimetic models will be able to generalize more accurately from opponent's past game play than humans. Indeed, the value of machine learning in many settings is that it can detect patterns and signals that go overlooked by humans. Yet it is still an open empirical question if playing a mimetic model offers meaningful advantages over traditional training methods.

 %We expect that mimetic models of specific participants would make this problem worse, by allowing for practice against individuals instead of styles or general trends.

If it turns out that mimetic models enhance a person's ability to prepare to play an opponent, then mimetic models have obvious implications for fair competition, especially if mimetic models are not universally available. We might be less concerned with such a development if the opponent that the person is preparing to play also had a mimetic model of the person to train against. But if only one of the two opponents has access to a mimetic model, then it poses an obvious threat to competition. While certain chess players might already benefit from access to resources and training that are not available to others, mimetic models could further exacerbate these disparities, eroding the equal playing field on which we hope players will compete.

%\eissue{Gaining Advantage Over the Target} The core issue raised here is that the use of \textit{mimetic models} will allow for a larger advantage than simply possessing the data used to create them, i.e. we assume that the person interacting with the model (\textit{Interactor}) is somehow more efficiently or more effectively learning to generalize from past actions via interactions with the model. While the pool of uses where this is true right now may be small: \say{\textit{attacks only get better with time}}~\cite{garman2015attacks}, so we can expect the pool to grow.

%\eissue{Deterioration of Competition} A longer term issue that the use of \textit{mimetic models} may cause is the degradation of the optimal preparation strategy for competitions. 

\xhdr{Preparing for an interview} Consider a person about to undertake a job interview who happens to have access to a mimetic model of the person who will interview them. The interviewee might attempt to gain an edge on the interviewer by completing a round of test interviews with the mimetic model. In so doing, the interviewee might learn the specific things about themself that they would be wise to withhold and the specific things about themself that they would do well to highlight---that is, the interviewee might be able to figure out how to make the best possible impression, given what they have to offer and given what the interviewer is looking for. Access to a mimetic model of the interviewer could also allow the interviewee to test out different persuasive styles. Even when presenting the exact same facts about themself and their career, the interviewee might communicate these quite differently, with some presentations of these facts being much more compelling than others from the point of view of the interviewer. The interviewee might therefore test out a range of different approaches on the mimetic model, adopting a more aggressive and boastful style in one interaction before trying out a more agreeable and modest style in the next. The mimetic model could help the interviewee hone their tone to increase the likelihood that the interviewer will be left with a favorable impression. The interviewee could even rely on the mimetic model to learn personal details about the interviewer that would seem to have nothing to do with the job, but which might help the interviewee cultivate greater rapport with the interviewer. For example, the interviewee might learn that the interviewer is a baseball fan, that they own two dogs, and that they had a difficult divorce. The interviewee might try to establish some degree of affinity with the interviewer by strategically weaving these topics into the conversation, bonding over shared interests and gaining confidence by demonstrating sympathies for personal challenges. 

Note that this scenario differs from the previous one insofar as the interaction is not zero-sum. In chess and other competitions, one person's gain is another person's loss: when a person learns the weakness of their opponent, the opponent necessarily suffers. The situation is different in the case of a job interview because there can be some alignment of interests. An interviewer might be pleased that the interviewee has communicated information about the characteristics of interest. Setting aside the possibility that an interviewee might simply lie about their qualifications or manufacture details that their interactions with the mimetic model suggest would impress the interviewer, there can be mutual benefits to an interviewee learning how best to interact with an interviewer. Of course, many of the things that the interviewee might learn about the interviewer via the mimetic model might be valuable not because they allow the interviewee to be assessed more accurately on their merits. Instead, the mimetic model might reveal personal qualities about the interviewer that the interviewee can exploit to compensate for their lack of merit. It's not obvious that an interviewer would be well served by someone who has simply figured out how to push their buttons. 

Indeed, mimetic models could easily make people far more vulnerable to manipulation and exploitation. In everyday life, people rarely have the chance to try their luck multiple times to figure out the optimal steps to get what they want from an interaction. Learning intimate details about a person---their preferences and propensities, but also deeply private facts---often requires making yourself vulnerable to the person in the process. Their is some risk involved in feeling out an interviewer: they get to know something about you as you try to get to know something about them. Mimetic models undermine this symmetry. 

\xhdr{Beyond interviews}
While we've focused on interviews, such dynamics apply to a range of activities in which two parties are attempting to learn about and assess each other. As mentioned earlier, a mimetic model might help prepare for pitch meetings, but also interactions that seem much more distant from interviews. 

Dating is a particularly useful scenario to contemplate because our instinctive reactions to using mimetic models in that context are normatively instructive. 
Imagine that $A$ is going on a first date with $B$ and hopes that it will lead to a longer-term relationship;
and imagine that, as in our job-interview scenario, $A$ prepares for the date by interacting with a mimetic model of $B$.
There are some basic contrasts with the job-interview setting that may shift our normative assessment.
In particular, a job interview is fundamentally transactional, and we evaluate the use of a mimetic model against the integrity of the transaction.
In contrast, a first date is part of a potentially longer-term relationship that involves a range of other qualities, including establishing trust as a basis for intimacy, and the way in which this trust is established through expectations about the nature of the interaction.

We can also ask how the use of a mimetic model differs from other forms of preparation that $A$ might do for their date with $B$, such as asking $B$'s friend $C$ for advice on what to emphasize in conversation.
We have an intuitive sense that interaction with an actual model of $B$ may be a qualitatively different type of preparation; this difference is reflected in pop culture's fascination with versions of this precise scenario, in the perfecting of repeated interactions in movies like {\em Groundhog Day}.
Indeed, to have access to a mimetic model of someone begins to approximate the experience of being able to repeat a ``time loop'' with them.
And this is a reflection of a point from earlier in this section, that the power of machine learning in general is to identify patterns that escape the unaided perception of human beings. In this way, the mimetic model of $B$ may encode things about $B$ that would be practically infeasible for $A$ to discern on their own. 

% Reid, I think there is a nice way to connect consent to the above sentence. Something like "Indeed, the dating scenario makes salient the lack of consent...", or something like that. I would note that even if we assume public data, the superhuman ability of mimetic models to extrapolate raises a concern about whether consent is warranted. Then you can end with that sentence about superhuman assistance below. 

The dating scenario makes salient the lack of informed consent~\cite{department_of_health_education_and_welfare_belmont_2014}. In this example and others, the mimetic-model-informed interaction is made more powerful by the target's lack of knowledge of how the model was used. Even if the model were trained on purely public data that the target was aware of, the model's (potentially superhuman) ability to provide specialized feedback in concrete situations raises a natural concern about whether this use requires consent. As the creator of a mimetic model often differs from the target, the question of consent persists through all of the scenarios we consider.    %This superhuman level of assistance in navigating social interactions may be a central reason why the use of such assistance seems to fundamentally shift our basis for establishing trust from interpersonal interactions.

\subsection{Mimetic Modelling as an End in Itself}\label{sec:ends}

% Counter factual
% Providing access/audience
% Population of worlds
% Stand in / substitute at work

In addition to being used indirectly to inform some future interaction, mimetic models could also be used directly as ends in their own right. In this Section, we explore a number of scenarios in which one's interaction with mimetic models is the end goal. 

%I also agree on the distinction in "Ends" between interactions and outputs. Certainly spectating is a reasonable "end", and distinct from directly interacting with a model as an end. I think having work products distinct from interactions makes sense as well -- if we want to distinguish them from the "deep fake" line of work in which the concern is about a single object, I think one point is that our scenarios involving work products include the ability to observe a continuous stream of new work products as the model reacts to new situations. This still is distinct from the operator directly interacting with the model; instead, they're able to reliably observe the model in a sequence of new situations and see how they respond.

\xhdr{Target replacement} 
In many cases, people will be able to deploy mimetic models directly into important interactions. For example, imagine that an entrepreneur runs an online tutoring service, and employs a particularly popular and idiosyncratic tutor $A$. When parents inquire about the tutoring service, they most often wonder if $A$ is available to teach their children. If the entrepreneur has access to a mimetic model of $A$, they could temporarily substitute the model for $A$ when $A$ is unable to work, for example if $A$ is out sick. If the mimetic model satisfies customers just as well as $A$ does, the entrepreneur may wonder if they still need $A$'s services at all, and could opt to permanently replace $A$ with the mimetic model of $A$. The entrepreneur may even go further, and wonder if the customer base as a whole would be more satisfied if everyone could be served by the model of $A$, rather than the various other human tutors under their employment. As another example, imagine that the reigning chess world champion Magnus Carlsen is not available to play in the online tournament you are organizing. You could opt to substitute the mimetic model of Magnus so that the other participants and viewers get to experience playing with and watching a proxy of him. 

As a related scenario, whenever a mimetic model is available, there is the possibility that people will use it to have a ``private audience'' with a simulated version of the target. The age-old question ``If you could have a conversation with any person, living or dead, who would it be?'' may not be so hypothetical with mimetic models. Given access to the appropriate model, one could talk with a proxy of a famous world leader, a respected author, or a celebrity. 

These scenarios raise the clear threat of targets being devalued, or even replaced, by their respective mimetic models. If interactors enjoy interacting with the model of $A$ as much as---or more than---interacting with $A$ themself, then $A$'s position in social and economic marketplaces is compromised. In the more extreme versions of this scenario presented above, $A$'s work could even be completely replaced by the work produced by $A$'s model. It is important to note that this raises a new question for the future of work, as $A$'s replacement is valuable because of $A$'s unique qualities, which a mimetic model can capture but a traditional ML model cannot. In contrast, most of the discussion around automation and human labor has focused on situations in which humans performing generic tasks are replaced with generic machines. Here, individual people who currently have no substitutes at all, human or machine, are now threatened with the prospect of mimetic models that can partially or completely substitute for them. Chess champions such as Magnus Carlsen have traditionally commanded up to tens of thousands of dollars for the chance to play them in a single game. Similarly, top e-sports professionals are paid hefty appearance fees to participate in events. How might this change if mimetic versions of these players are widely available?

In addition to these labor considerations, people valued for unique traits, outputs, or interaction styles could find themselves devalued by the presence of mimetic models that capture their signature styles to a reasonable degree. If people are satisfied by having a private audience with mimetic proxies, the targets may consequently be less in demand. Individuals may lose some of their social capital if part of their uniqueness is lost to mimetic models. Perhaps even friends would be less in demand---if the mimetic version of your friend can do a convincing job of reacting to your stories or problems as they would, how will that affect your friendship? 

\xhdr{Mimetic counterfactuals} Mimetic models, by generating realistic actions faithful to a specific individual's style, could be used to play out various counterfactual scenarios. For example, fans of creative endeavors often speculate what would have happened if person X had been in situation Y.  For example, what would have happened if Bobby Fischer had shown up for his 1975 World Championship match with Anatoly Karpov instead of forfeiting it? How might Mozart's music have evolved if he had lived past 35? What did the letters that Nora Joyce wrote to her husband  James Joyce contain before their grandson burned them? In principle, one could employ mimetic models to attempt to answer these kinds of questions. A Fischer model and Karpov model could face off under 1975-like conditions to shed light on who might have won; a Mozart model that can extrapolate from his earlier styles to his later styles could further extrapolate beyond his death; a Nora Joyce model could ``respond'' to James Joyce's still-existing letters (and we might even judge the Nora model's attempt to fill in the gaps by how faithfully a James Joyce model's response adheres to his actual reply).

%How would Magnus Carlsen have navigated the chess position you eventually lost yesterday? 
Beyond historical questions, one could also explore contemporary counterfactuals via mimetic models. How would your idea for a song have turned out if you gave it to Taylor Swift? Which of your brilliant chess moves would the current World Champion Magnus Carlsen have failed to find? How might a debate between politicians go with the prompt you wish had been asked? Contemporary figures are just as easily modeled as historical ones, if not more easily due to the generally increased training data available. 

In all of these scenarios, perhaps the most immediate ethical implication is the risk of reputational damage to the targets. To the extent that the models are imperfect representations of their targets, they will occasionally deviate from what the target would actually do. These deviations, especially salient or problematic ones, could alter what others think of the target. And more generally, we cannot know how accurate a mimetic model's behaviors are at extrapolating to a fully counterfactual scenario. If the mimetic simulation of the 1975 World Championship ends up with Karpov dethroning Fischer, that could alter the public's perception of these two players. If the Mozart model ends up reproducing musical innovations that others later conceived, the credit for them may shift. If the Nora Joyce model outputs offensive content, historians may think of her differently.  We are familiar, for example, with similar effects arising from inaccurate public perception of real events based on historical fiction, such as when obituaries of Mark Felt (who served as Bernstein and Woodward's anonymous source in the Watergate scandal) attributed the quote ``Follow the money'' to him, despite the fact that this quote was uttered only by his fictional counterpart in William Goldman's screenplay for the movie {\em All the President's Men} \cite{greenberg2018william}.

Importantly, the risk of reputational damage in these counterfactual scenarios could actually \emph{increase} with the accuracy of the models. If mimetic models aren't accurate, people will be less likely to trust them. A mimetic model that makes obvious or frequent mistakes would come across more as a caricature than a realistic representation. If one's expectations of the model are low, then mistakes, deviations, or questionable outputs could easily be attributed to quirks of the model rather than traits of the target. But if highly accurate mimetic models, such as those that already exist in chess and writing, were to generate the same mistakes or questionable outputs, they could be interpreted very differently. An accurate mimetic model engenders trust by generating realistic outputs, including ones we can validate by comparing with the target's actual response to the same input situations. Whatever outputs they generate will typically be treated as more reflective of the target rather than model artifacts. 

In addition to the reputational damage that individual targets may suffer, mimetic models may be systematically biased in their misrepresentations. As a result, entire populations of people may be perceived worse because of how they are mischaracterized by mimetic models. Again, this risk is pronounced for generally accurate models that engender more trust by end-users. Although many ML models have been found to be systematically biased against particular subgroups, algorithmic bias that arises in mimetic models could pose new risks. Since mimetic models differ from each other by definition---as they target different individuals---systematic errors across a particular subgroup could be mistakenly attributed to the subgroup rather than arising from correlated flaws across many different models.

\subsection{Case Study: Mimetic Models of Oneself}\label{sec:self}

An interesting case arises when considering the use of a mimetic model where the creator, operator, and target are the same person---in other words, when an individual creates and deploys a mimetic model of themself. Such a mimetic model may be used both as a means to an end and as an end in itself, traversing the scenarios discussed above and their associated ethical and social considerations.

%Most of these uses are for the use as a means, but their are uses as end available too, e.g. you want to play a version of yourself in a video game. But we will focus on the use as a means as that is where the unique ethical issues are presented.

%We consider the use of \textit{mimetic models} where the \textit{operator} and \textit{target} are the same person to be a special case requiring separate study. Most of these uses are for the use as a means, but their are uses as end available too, e.g. you want to play a version of yourself in a video game. But we will focus on the use as a means as that is where the unique ethical issues are presented.

Consider the use of a mimetic model of oneself as an end. One natural use case for such a model is as a stand-in for work: a mimetic model can perform work on a person's behalf without them having to expend any effort or time. For example, a person might create a mimetic model that predicts their own responses to messages~\cite{Zhang2018Jan}, such as e-mail from work colleagues, and sends responses automatically on their behalf~\cite{jahanshahi2021auto}. By creating and operating multiple models, the person can essentially use mimetic models as a {\em force multiplier}, to scale out their work and increase the number of people they interact with. For example, an artist specializing in portraiture could use a mimetic model to create portraits of customers, given a photograph, in a style that mimics what they would have created by hand~\cite{xu2022deep,ci2018user}. This would enable the artist to create many more custom portraits than would be physically possible.

In a similar vein, a mimetic model could enable someone to provide a private audience for multiple people at the same time. For instance, a sought-after chess coach could interact with multiple students at the same time by having them play against a mimetic model that captures the coach's playing style and decisions~\cite{mcilroy2020learning}, providing each student with a private, one-on-one training experience. Although the coach could alternatively play an online simultaneous exhibition against the students, rotating through the games and making each move, this would be physically and mentally taxing for the coach, and the quality of each game would degrade as more students are added. In contrast, a mimetic model of the coach would not be subject to these physical limitations.  

A mimetic model of oneself could also be used as a means to an end. One natural use case is to allow a mimetic model to interact with other people or entities before interacting with them in real life, as a way of filtering or preparing for these interactions. For example, a person who wishes to join an online dating site may be unfamiliar with the site's population or environment~\cite{ranzini2017love}. By creating a mimetic model of themself and allowing it to interact with the online site and its participants, they can observe the outcomes of these interactions and selectively pursue the interactions that seem most promising in real life. 

While hypothetical, several of the above scenarios are within reach today. Large-scale language models have shown great promise in being fine-tuned to specific applications~\cite{codex}; fine-tuning them to an individual's writing style is within reach. Personalized models of chess can already be trained with high enough accuracy to uniquely identify each player, given a moderate number of games per player~\cite{mcilroy2020learning}. And while artists, musicians, and authors have long used ``ghost'' assistants to scale out their work, the rise of mimetic models is bringing an unprecedented automation to this practice. 

%The proliferation of mimetic models, tailored to specific tasks or scenarios, is becoming a potential reality we must face.

In all of these scenarios, the target, creator, and operator of the mimetic model are the same person. This presents a different subtlety to the ethical issues raised in previous sections, because issues of privacy or consent in the creation and use of the mimetic model diminish---the target of the model, being the individual themself, already embodies these rights---whereas issues of disclosure, value, and impact become more prominent. To start with, what level of disclosure is appropriate for the authorship of the mimetic model's communication and actions~\cite{simmons2020catfishing}? Should each e-mail message written by a mimetic model be explicitly flagged as such, so the recipient knows it was not written by a real person? What is the monetary value of artwork created by a mimetic model compared to artwork created by a real person? How do our answers change if the output produced by the mimetic model is perceived as better than what the target individual would have produced, or worse?  

%The latter is possible in many domains, since humans are not perfect and often take suboptimal actions or generate suboptimal content; in domains where superhuman AI exists, like chess, this is always possible, regardless of how strong the human is.

These questions underscore the importance of fidelity as a dimension for assessing the value of a mimetic model. If a mimetic model does a poor job of mimicking the target individual (i.e., it has low fidelity), then its value is clearly decreased and interactors will reject the model's similitude to the target. A more interesting situation arises when the mimetic model has high fidelity. In this case, even if interactions with the model faithfully simulate interactions with the real person, a person using multiple mimetic models of themself might potentially reduce the value of each interaction. Does a ``thank you'' e-mail sent by a mimetic model, however authentically crafted, evoke the same level of gratitude as a message written by the actual person? Should original artwork generated by a mimetic model command the same price as original artwork created by the actual person? Could a practice chess game with a mimetic model of a coach provide a better learning experience than a real game with the coach, if the coach is distracted or tired in real life? In all of these situations, the distinction between mimetic output and real output, and the relative quality of these outputs, influences the value that interactors will attribute to the corresponding interaction.
%No matter how physically identical cubic zirconia is to a real diamond, it will never command the same price. 

Note that the devaluation mentioned above may constitute an acceptable trade-off for a person: even if interactions with their mimetic models are valued less than interactions with the person in real life, the scalability of mimetic interactions could make them financially advantageous to the individual. For example, a chess coach might provide a discount for playing training games with their mimetic model (and support thousands of students simultaneously), while charging substantially more for playing with them in real life. %\sid{Question for Ashton: should we just cut the rest of this paragraph from "Taken ..." onwards?} Taken to the extreme, excessive proliferation of mimetic models could lead to catastrophic devaluation of an individual's services---essentially, by flooding the market with excess supply---and could lead to a general distrust of online personas~\cite{skjuve2019help}. This, in turn, may create a market for large-scale Turing Tests~\cite{imitationgame} of some form to help distinguish between real vs. mimetic interactions. 

An interesting ethical consideration arises when a mimetic model of a person behaves differently than the person would, whether in a positive or negative sense. As a positive example, consider a mimetic model that responds to email using a level of politeness that is higher than the target individual's natural politeness. The responses may be adjusted to avoid language that some readers might find offensive; indeed,
%Similarly, a mimetic model of a chess player may play at a higher rating than the target player, while the results and interactions of the mimetic model may still be attributed to the target player. the results of its interactions are attributed to the real target player.  
one can imagine a marketplace of apps that filter or modulate a mimetic model's output to achieve desirable properties. Such intentional modifications to a mimetic model's output could raise ethical considerations because they misrepresent the target individual and may be viewed as deceptive. 

Mimetic models may also deviate from the target individual's behavior in a negative sense, for example by exaggerating a negative tendency %or bias. This concern is not limited to mimetic models, and is the focus of a large body of work on safety and fairness in machine learning~\cite{Amodei2016ConcretePI,Matthews2022TheAP,Mehrabi2021ASO}.
However, since a mimetic model acts as a stand-in for the target individual, its actions have direct implications for the individual's reputation and their liability in the event of harms being inflicted on the model's interactors. These harms extend beyond ``noise'' in the model training process and include endogenous biases that exist within the individual themself, which may be adopted or even amplified by the mimetic model. If the mimetic model is deployed at scale, this could result in the individual's biases being proliferated at scale. For example, if an individual who is prone to offensive comments creates and deploys mimetic models of themself on various online dating sites, this could amplify the effect of such individuals on these sites. 

%% file: 5-ethical-discussion.tex
\section{Overview of Ethical Themes}\label{sec:disc}

%\sid{I don't think I'd title this "Ethical Issues in Detail", since we are providing a high-level summary here. Perhaps "Ethical Themes" or "Summary of Ethical Issues"?"}
%\jon{I agree, and made a proposed change of the section title to "Overview of Ethical Themes"}

Having examined the ethical questions that arise when mimetic models are deployed in a range of specific scenarios, we now discuss some of the common themes that run through these scenarios, and their implications more generally.

Several themes recur in our analyses. First, the presence of mimetic models has the potential to significantly alter the relationships between people across a variety of settings. One of the simplest but clearest demonstrations of this is in competition: unequal access to mimetic models could substantially change the nature of who can compete, and the outcomes that can arise from competitions. This holds for both models used as a means to an end---e.g., in preparing for upcoming competitions---and as an end in themselves---e.g., in replacing real people with mimetic models of them. Relatedly, mimetic models have the potential to seriously change how individual people are valued. To the extent that individuals are valued in certain settings for their idiosyncratic behaviors and products, either socially, economically, or otherwise, and to the extent that mimetic models can faithfully simulate these behaviors, there may be significant effects on how people are valued. This also includes concepts of self-worth: how people value themselves could be influenced by how the interactions and outputs of their mimetic models are valued. An interesting consideration for a more distant future is how the value of human-ness itself might change in a world where mimetic models are powerful and commonplace. Will the role of friendship change if a mimetic model can fulfill some of the functions that human contact currently plays; or perhaps will in-person interactions with real people become more important, to guarantee that you are engaging with an actual person and not their mimetic model?

Another consistent theme across our scenarios is the increased capacity for bad-faith activities using mimetic models. Although we did not analyze deceptive practices in depth since they are already relatively common, mimetic models may make deception an even more prevalent threat. Imagine a phishing attack where a scammer pretends to be a trusted party, and can sustain a prolonged interaction posing as this trusted party. Mimetic models also increase the scope for manipulation. If one can thoroughly test how a particular target person will react to a wide variety of prompts or actions, it becomes more feasible to identify weaknesses that can be exploited for one's own benefit. Finally, the new privacy risks posed are easy to see. Mimetic models could qualitatively change our ability to process past behaviors and generalize to novel situations, thus raising the prospect of unintentionally leaking information about ourselves, our behaviors, and our identities. 

%\sid{I hesitate to introduce a third dimension here ("modality"), which we have not previously discussed. I think we can just merge this with "generality", which we do previously mention (though arguably do not discuss enough).}

%\jon{This last paragraph is very similar to the last paragraph just before the start of Section 2.1.  Since this version of the paragraph is a bit more extensive, and arguably in a better location in the paper for these points, I went back and commented out the paragraph just before the start of Section 2.1.  I'm okay with leaving in "modality" since --- although we haven't really used the term --- it's implicit in the way we move between examples from game-playing, text-generation, and other styles of output.}

%\sid{Yes, this works! My main concern was the inconsistent repetition relative to what was in section 2.0 (I'm assuming that's what you meant). Since you removed that and put it here, this concern goes away.}

Finally, we take note of three important dimensions of mimetic models that appear to play an influential role in determining the ethical consequences their use may have. First is the \emph{fidelity} of the model, or how faithfully it captures its target's behaviors and characteristics. Many of the ethical issues we have discussed become more salient as model fidelity increases. If a mimetic model is only passably accurate, and is often easily distinguishable from the target, then it becomes more of a caricature than a realistic simulation. As such, issues such as deceptive practices and reputational damage become less of a concern. Second is the \emph{modality} of the model, the domain it operates in and the types of behaviors it is designed to reflect. Clearly, a model that can output text differs from one that can output chess moves, and the ethical issues raised by each differs as a result. Third is the \emph{generality} of the model, or the breadth of scenarios and domains that a mimetic model can capture. Generally speaking, the wider the model's reach, the more pertinent the ethical concerns.

%% file: 6-related-works.tex
\section{Related Work}

\subsection{General Considerations}

Some of the initial discussions of mimetic models occurred in science fiction (e.g. \cite{brin_2002,egan_2002,pohl1977gateway,vinge_monster}), 
but our understanding of them has become much more specific as the technology to produce them has become concrete and increasingly available.
\nocite{taddeo2018ai,fish2021reflexive,whittaker2019disability,guo2021detecting,fazelpour2020algorithmic,mehrabi2021survey}
Our discussion of the normative considerations related to mimetic models in turn connects to some of the central themes in the ethics of AI, including the fairness of decisions~\cite{selbst2019fairness,barocas2017fairness,finocchiaro2021bridging,chouldechova2020snapshot}, the potential for bias~\cite{raghavan2020mitigating,buolamwini2018gender,harcourt2015risk,kleinberg2018discrimination}, and potential shifts in accountability~\cite{kleinberg2018human}. 

Mimetic models also introduce questions related to  data access~\cite{zwitter2014big} and informed consent~\cite{department_of_health_education_and_welfare_belmont_2014}, and may benefit from strategies such as  \textit{Model Cards}~\cite{mitchell2019model} to address these issues. When mimetic models are produced on anonymized data, they introduce the risk of deanonymization through their behavior, based on some of the principles in the privacy literature \cite{narayanan2008robust,zimmer2020but}.
Mimetic models contain significant potential for deception as well, and the issues here are related to the issues that arise with {\em deepfakes}~\cite{vaccari2020deepfakes}, as we discuss next. Some of the concerns associated with this type of deception are fake announcements by public figures~\cite{agarwal2019protecting}, devaluing of performers~\cite{Rosner2021Jul}, and fake news~\cite{zellers2019defending}.

%No explicit discussion, limited discussion in some papers. Related concepts do have good ethical discussions. Additionally their are many ethical issues that are not unique to mimetic models: their potential use for deception is high but as people have been misleading other people for millennia their is good ethics discussion on pure deception so we will focus on ethical concerns that are specific to mimetic models instead of all those that could be relevant. Additional issues that could be relevant are but limited to: deanonymization, bias, misuse of data, lack of informed consent, ...

\subsection{Related Concepts}\label{sec:related}

As noted in the introduction, it is useful to explore the relationship between mimetic models and related concepts at the boundary of AI modeling and human behavior. 
We consider a number of these in this subsection.

\subsubsection{Deepfakes}\label{sec:deepfake}

Deepfakes raise normative concerns that overlap those encountered with mimetic models. The term \textit{deepfake} refers to a set of techniques for manipulating video or images to replace or generate the likeness of a person.\footnote{The techniques are not limited to humans, but we focus on their application to humans here.} 
The name originates from a deep-learning face-swapping program, popularized by the Reddit user \textit{deepfakes}, that allows a user to replace the face of an actor in a video (or still image) with that of another target~\cite{deepfake, TOLOSANA2020131}. Importantly, the requirements for training the model are low, the system can be run on a single consumer-level GPU, the replaced video can be low resolution, and the number of samples required for the target can be as little as a single image.\footnote{More angles/lighting conditions lead to a better result, so multiple images are required to generate a more dynamic set of outputs.} 

Expanding beyond this specific origin, the term \textit{deepfake} has grown to acquire a broader definition in the culture more generally (e.g. \cite{deepfake_report}), and is now viewed as a key component in \textit{fake news}~\cite{lazer2018science,Korshunov2022}.  As noted in the introduction, a key distinction between even this broader framing of deepfakes and the concept of a mimetic model is the fact that mimetic models are designed for interaction in new  situations.  We require mimetic models to be able to interact with people, in which they take some action, observe the response, and take another action based on the response. In contrast, deepfakes are typically pre-generated for a single planned behavior.
%\footnote{The most common is an actor for the body language/visuals, with a separately generated audio track.} \sid{Not sure how useful this footnote is, and it's a bit difficult to parse; consider cutting.}
One way to think of the relationship is to note that a mimetic model could naturally be used to generate the text spoken by a deepfake model.
Of course, the distinction is not absolute, and adding interactivity to a deepfake would produce a type of mimetic model. 

\subsubsection{Digital Avatars}

Many people employ visualization of their online persona that is distinct from their own physical body, be it a simple cartoon image or a complex 3D model~\cite{Ducheneaut2009BodyAM,Suh2011WhatIY}. These avatars act on behalf of the "target", to use our framework's terminology, either directly under the control of the target or in some pre-programmed way. Thus the concerns that misuse or mistreatment~\cite{Dechant2021HowAC, Hill2013AvatarEB} of avatars raise has overlap with those of mimetic models. Additionally, people can become attached to their avatars both emotionally~\cite{Wolfendale2006MyAM, Fox2009VirtualST} and through their physical representation~\cite{Yee2007ThePE,Peck2013PuttingYI}; having a virtual representation of yourself can in some cases lead to a phenomenon known as the \textit{Proteus effect}~\cite{Yee2009ThePE,Yee2009ImplicationsOT}, in which people adapt their behavior based on characteristics of the avatar.  The use of avatars to test new experiences overlaps with the use of mimetic models as proxies, as we discuss in Section~\ref{sec:ends}.

\subsubsection{Style Transfer}

Style transfer \cite{Gatys2016, Li2017Jan, Krishnan2021Jun} is a technique in which an algorithm transforms a piece of media to render it in the style of a specified target author.
\nocite{park2019gaugan}
Style transfer techniques typically use a single static initial image~\cite{Gatys2016}, video~\cite{sanakoyeu2018style}, audio clip~\cite{cifka2020groove2groove}, or other representation \cite{ma2019neural}.
However, at a broader level of abstraction, they can be viewed as creating a special-purpose mimetic model of the target author, for the purpose of interacting with a prompt to produce new work in the target author's style.

%\sid{Consider merging the multi-modal section with the style transfer section?}
%\jon{Given that the multi-modal section is about crossing between modalities, which is a new point, and given that each of these subsections seems to be only a paragraph, it seems consistent to keep them separate.  But I'm fine either way.}
%\sid{Sounds good}

\subsubsection{Multi-modal generative agents}

There has been ongoing progress in machine learning systems that translate prompts such as `A Mayan warrior getting ready, in the style of Rembrandt'~\cite{dalletweet} into an image matching the prompt---e.g., ImageBERT~\cite{Qi2020ImageBERTCP}, ALIGN~\cite{Jia2021ScalingUV}, CLIP~\cite{radford2021learning} and DALL$\cdot$E~2~\cite{Ramesh2021ZeroShotTG,ramesh2022hierarchical}, or the reverse (images to text) like Flamingo~\cite{Alayrac2022FlamingoAV}. These systems allow for outputs that mimic the styles of specific individuals, and can be fine-tuned to allow for style transfer~\cite{Liu2022NameYS}. Generating mimetic models is not the main goal of these works, but they may be the foundation for mimetic models.

\subsubsection{Model Personalization}

Personalized systems are those that adapt their outputs  to the user they are interacting with~\cite{mcauley2022personalized}.
This is often done by creating a model that interacts with a user over time, maintaining and improving a representation of the model's knowledge about the person~\cite{kang2018self,wu2019session}. The task is thus a type of mirror image to what a mimetic model does: personalization seeks to create a model that can make optimal responses to a user, while a mimetic model instead seeks to act as a stand-in for the user and generate responses in their stead.

\subsubsection{Legal Stand Ins}

A non-computational analogy to mimetic models in the off-line world can be found in the way that legal systems allow for proxies~\cite{easterbrook1983voting}, power-of-attorney, or other mechanisms to allow a designated individual to make decisions that are intended to represent the intent of a specific target person. As a result, the history of ethical considerations involving proxies can provide insights into the corresponding issues that may arise with mimetic models~\cite{rezaee2008corporate}.

\subsubsection{Other Concepts}

Finally, we touch on a few additional concepts more briefly.

%\xhdr{deepfakes}, these are systems that generate images/videos/audio of people, the technology has many ethical considerations that are outside of the scope of this paper as the models do not interact with the world. They are merely puppets whose decisions are controlled. We see the difference between deepfakes and mimetic models to be two sides of a continuum. With the amount of decision making starting at 0 with simple deepfakes, moving to one style transfer where partial decisions are made (i.e. a prompt is provided). Then to fully mimetic systems where the prompt is minor or non-existent. 

\xhdrsub{Recommender Systems} Systems that recommend content by modeling a user's preferences  ~\cite{bobadilla2013recommender,resnick1997recommender} are not mimetic in our sense, since they are not generating behavior on behalf of the user.  However, we can imagine ways in which mimetic modeling ideas could be incorporated into a larger recommendation context, such as through mimetic modeling of the next movie selected to play (i.e. autoplay behavior)~\cite{jenner2016tviv}.% \sid{Not sure what you mean by "autoplay" here -- do you mean video/song autoplay?}

\xhdrsub{Work Automation} There is of course a vast literature on automation, and the ways in which AI in particular is replacing certain categories of jobs~\cite{Acemoglu2019Automation}. 
Our analysis overlaps with this literature only to the extent that jobs are being replaced by models of {\em specific} workers, rather than the typical practice of designing AI or ML systems to perform well on the underlying task in a generic or aggregate sense.
This distinction also applies in the context of automation via robotics \cite{Acemoglu2020Robots,kober2013reinforcement}.

\xhdrsub{Prediction}  There are well-established methodologies for converting a generative system to a predictive one~\cite{ng2001discriminative}, and via this principle mimetic models can be used to predict a person's behavior, simply by observing the behavior that is generated by the model.  This translation implies that mimetic models share the same concerns about predicting the behavior of individuals~\cite{crawford2014big}. 

\xhdrsub{Speculative Fiction} As noted at the start of this section, many of the ethical issues we discuss here are also found in works of fiction~\cite{mccarthy_2003,egan_2010}. Fictional approaches to these questions are not bounded by real-world constraints, and so they are often much more exaggerated in their formulations than what we consider here.  For example, works like David Brin's \textit{Kiln People}~\cite{brin_2002}, Greg Egan's \textit{Zendegi}~\cite{egan_2002} or Vernor Vinge's \textit{The Cookie Monster}~\cite{vinge_monster} all directly discuss the implications of high-fidelity models of specific people and their ethical implications.

\subsection{Current Applications of Mimetic Models}

One of the main realized uses of mimetic models in practice today is for game-playing.
As a general domain, games provide both highly detailed behavioral data~\cite{hooshyar2018data,Melhart} and easy creation of computer-controlled players~\cite{brockman2016openai}. There are also financial incentives for game vendors to provide mimetic models as a feature for users~\cite{Thomas2010Nov,Morris2013Aug,Tantaros2019Jan,gitlin2021Dec}. Chess~\cite{mcilroy2020aligning}, Go~\cite{silver2016mastering} and other~\cite{jacob2021modeling} tabletop games have also been studied in the context of creating  human-like models, usually with a focus on human-compatible agents~\cite{hu2021off} or creating tools for teaching humans~\cite{mcilroy2020learning}. 

Mimetic models have also been investigated in educational settings, with the creation of models of both student~\cite{Geigle2017Apr} and teacher~\cite{chaturvedi2014predicting} behavior. In these cases, however, the generative nature of the models was not the focus of the research. 
Content-filling algorithms such as in-painting brushes~\cite{guzdial2019friend,Koch2019MayAD} can also be viewed as a type of mimetic model, raising similar issues to applications in text generation discussed in Section~\ref{sec:self}. 
Finally, mimetic models have been used to encode individual artistic style; one example is in archaeological studies of pottery~\cite{ramazzotti2018encoding}, where the goal is to generate similar pieces of pottery based on the styles of specific artisans, or models like DALL$\cdot$E~2~\cite{ramesh2022hierarchical} that can create an image in the style of a specific artist matching a prompt. 
%\sid{I could be wrong, but I feel like we are missing other examples that we have previously described as being mimetic. Perhaps those examples (like producing content in the style of an artist/composer) are not explicitly generative models, and hence we are not including them here?}

%Main current uses are: games, both as opponents and for understanding players, teaching, auto complete systems, counterfactual explorations, and as byproducts of powerful ML systems (GPT-3/CLIP) 
%There are many current systems that marginal, this is byproduct of the newness of mimetic models~(). 

%% file: 7-conclusion.tex
\section{Conclusion}

Mimetic models represent a complex new direction in the use of AI to model human behavior---one in which models are tailored to match the behavior of specific individuals, and in settings that allow for rich interaction with others. We have seen that mimetic models surface subtle ethical and social considerations across a wide range of scenarios---including as forms of preparation for future interactions with real people (in a competition, an interview, or a date); as an end in themselves to study counterfactuals or to provide spectator experiences that would be hard to produce using real people; and as a way for people to create realistic stand-ins for themselves. We believe that the framework here suggests a number of directions for further investigation, including more extensive domain-specific considerations as more powerful mimetic models become available across an increasingly wide array of contexts. 

% Mention EU AI Act

%\xhdr{Forward Visions} While it is beyond the scope of this paper to posit specific

% Work on new AI models continues to reveal new uses---and new issues. We are starting to see a new type of ML system \textit{mimetic models}, these are learned models of \textit{specific} individual's decisions. These models have many potential uses: algorithmic aids to teaching, allowing for richer virtual worlds, or allowing people to automate parts of their lives. These benefits come with trade offs, the models can be used to deanonymize students, lead to harassment of people based on their model's actions, or devalue the people you communicate with. This work seeks to act as first step in identifying these issues and allowing researches to start considering the implications of their work. 

%% file: 0-main.bbl
\begin{thebibliography}{100}

\bibitem{Acemoglu2019Automation}
{\sc Acemoglu, D., and Restrepo, P.}
\newblock Automation and new tasks: How technology displaces and reinstates
  labor.
\newblock {\em Journal of Economic Perspectives 33}, 2 (May 2019), 3--30.

\bibitem{Acemoglu2020Robots}
{\sc Acemoglu, D., and Restrepo, P.}
\newblock Robots and jobs: Evidence from us labor markets.
\newblock {\em Journal of Political Economy 128}, 6 (2020), 2188--2244.

\bibitem{agarwal2019protecting}
{\sc Agarwal, S., Farid, H., Gu, Y., He, M., Nagano, K., and Li, H.}
\newblock Protecting world leaders against deep fakes.
\newblock In {\em CVPR workshops\/} (2019), vol.~1.

\bibitem{deepfake_report}
{\sc Ajder, H., Patrini, G., Cavalli, F., and Cullen, L.}
\newblock The state of deepfakes landscape, threats, and impact, 2019.

\bibitem{Alayrac2022FlamingoAV}
{\sc Alayrac, J.-B., Donahue, J., Luc, P., Miech, A., Barr, I., Hasson, Y.,
  Lenc, K., Mensch, A., Millican, K., Reynolds, M., Ring, R., Rutherford, E.,
  Cabi, S., Han, T., Gong, Z., Samangooei, S., Monteiro, M., Menick, J.,
  Borgeaud, S., Brock, A., Nematzadeh, A., Sharifzadeh, S., Binkowski, M.,
  Barreira, R., Vinyals, O., Zisserman, A., and Simonyan, K.}
\newblock Flamingo: a visual language model for few-shot learning.
\newblock {\em ArXiv abs/2204.14198\/} (2022).

\bibitem{bard2020hanabi}
{\sc Bard, N., Foerster, J.~N., Chandar, S., Burch, N., Lanctot, M., Song,
  H.~F., Parisotto, E., Dumoulin, V., Moitra, S., Hughes, E., et~al.}
\newblock The hanabi challenge: A new frontier for ai research.
\newblock {\em Artificial Intelligence 280\/} (2020), 103216.

\bibitem{barocas2017fairness}
{\sc Barocas, S., Hardt, M., and Narayanan, A.}
\newblock Fairness in machine learning.
\newblock {\em NeurIPS tutorial 1\/} (2017), 2.

\bibitem{bird2020lstm}
{\sc Bird, J.~J., Faria, D.~R., Ek{\'a}rt, A., Premebida, C., and Ayrosa,
  P.~P.}
\newblock Lstm and gpt-2 synthetic speech transfer learning for speaker
  recognition to overcome data scarcity.
\newblock {\em arXiv preprint arXiv:2007.00659\/} (2020).

\bibitem{bobadilla2013recommender}
{\sc Bobadilla, J., Ortega, F., Hernando, A., and Guti{\'e}rrez, A.}
\newblock Recommender systems survey.
\newblock {\em Knowledge-based systems 46\/} (2013), 109--132.

\bibitem{brin_2002}
{\sc Brin, D.}
\newblock {\em Kiln People}.
\newblock Tor, 2002.

\bibitem{brockman2016openai}
{\sc Brockman, G., Cheung, V., Pettersson, L., Schneider, J., Schulman, J.,
  Tang, J., and Zaremba, W.}
\newblock Openai gym.
\newblock {\em arXiv preprint arXiv:1606.01540\/} (2016).

\bibitem{buolamwini2018gender}
{\sc Buolamwini, J., and Gebru, T.}
\newblock Gender shades: Intersectional accuracy disparities in commercial
  gender classification.
\newblock In {\em Conference on fairness, accountability and transparency\/}
  (2018), PMLR, pp.~77--91.

\bibitem{chaturvedi2014predicting}
{\sc Chaturvedi, S., Goldwasser, D., and Daum{\'e}~III, H.}
\newblock Predicting instructor’s intervention in mooc forums.
\newblock In {\em Proceedings of the 52nd Annual Meeting of the Association for
  Computational Linguistics (Volume 1: Long Papers)\/} (2014), pp.~1501--1511.

\bibitem{codex}
{\sc Chen, M., Tworek, J., Jun, H., Yuan, Q., de~Oliveira~Pinto, H.~P., Kaplan,
  J., Edwards, H., Burda, Y., Joseph, N., Brockman, G., Ray, A., Puri, R.,
  Krueger, G., Petrov, M., Khlaaf, H., Sastry, G., Mishkin, P., Chan, B., Gray,
  S., Ryder, N., Pavlov, M., Power, A., Kaiser, L., Bavarian, M., Winter, C.,
  Tillet, P., Such, F.~P., Cummings, D., Plappert, M., Chantzis, F., Barnes,
  E., Herbert{-}Voss, A., Guss, W.~H., Nichol, A., Paino, A., Tezak, N., Tang,
  J., Babuschkin, I., Balaji, S., Jain, S., Saunders, W., Hesse, C., Carr,
  A.~N., Leike, J., Achiam, J., Misra, V., Morikawa, E., Radford, A., Knight,
  M., Brundage, M., Murati, M., Mayer, K., Welinder, P., McGrew, B., Amodei,
  D., McCandlish, S., Sutskever, I., and Zaremba, W.}
\newblock Evaluating large language models trained on code.
\newblock {\em CoRR abs/2107.03374\/} (2021).

\bibitem{chopra2016meet}
{\sc Chopra, S., Gianforte, R., and Sholar, J.}
\newblock Meet percy: The cs 221 teaching assistant chatbot.
\newblock {\em ACM Transactions on Graphics 1}, 1 (2016), 1--8.

\bibitem{chouldechova2020snapshot}
{\sc Chouldechova, A., and Roth, A.}
\newblock A snapshot of the frontiers of fairness in machine learning.
\newblock {\em Communications of the ACM 63}, 5 (2020), 82--89.

\bibitem{ci2018user}
{\sc Ci, Y., Ma, X., Wang, Z., Li, H., and Luo, Z.}
\newblock User-guided deep anime line art colorization with conditional
  adversarial networks.
\newblock In {\em Proceedings of the 26th ACM international conference on
  Multimedia\/} (2018), pp.~1536--1544.

\bibitem{cifka2020groove2groove}
{\sc C{\'\i}fka, O., {\c{S}}im{\c{s}}ekli, U., and Richard, G.}
\newblock Groove2groove: One-shot music style transfer with supervision from
  synthetic data.
\newblock {\em IEEE/ACM Transactions on Audio, Speech, and Language Processing
  28\/} (2020), 2638--2650.

\bibitem{cimini2018walking}
{\sc Cimini, A.}
\newblock Walking to the gallery: {Sondra Perry’s “It’s in the game”}
  in {San Diego} in five fragments.
\newblock {\em Sound Studies 4}, 2 (2018), 178--200.

\bibitem{crawford2014big}
{\sc Crawford, K., and Schultz, J.}
\newblock Big data and due process: Toward a framework to redress predictive
  privacy harms.
\newblock {\em BCL Rev. 55\/} (2014), 93.

\bibitem{Dechant2021HowAC}
{\sc Dechant, M.~J., Birk, M.~V., Shiban, Y., Schnell, K., and Mandryk, R.~L.}
\newblock How avatar customization affects fear in a game-based digital
  exposure task for social anxiety.
\newblock {\em Proceedings of the ACM on Human-Computer Interaction 5\/}
  (2021), 1 -- 27.

\bibitem{department_of_health_education_and_welfare_belmont_2014}
{\sc {Department of Health, Education, and Welfare}, and {National Commission
  for the Protection of Human Subjects of Biomedical and Behavioral Research}}.
\newblock The belmont report. ethical principles and guidelines for the
  protection of human subjects of research.
\newblock {\em The Journal of the American College of Dentists 81}, 3 (2014),
  4--13.

\bibitem{dhou2018towards}
{\sc Dhou, K.}
\newblock Towards a better understanding of chess players’ personalities: A
  study using virtual chess players.
\newblock In {\em International Conference on Human-Computer Interaction\/}
  (2018), Springer, pp.~435--446.

\bibitem{Ducheneaut2009BodyAM}
{\sc Ducheneaut, N., Wen, M.-H., Yee, N., and Wadley, G.}
\newblock Body and mind: a study of avatar personalization in three virtual
  worlds.
\newblock {\em Proceedings of the SIGCHI Conference on Human Factors in
  Computing Systems\/} (2009).

\bibitem{easterbrook1983voting}
{\sc Easterbrook, F.~H., and Fischel, D.~R.}
\newblock Voting in corporate law.
\newblock {\em The journal of Law and Economics 26}, 2 (1983), 395--427.

\bibitem{egan_2002}
{\sc Egan, G.}
\newblock {\em Diaspora}.
\newblock Gollancz, 2002.

\bibitem{egan_2010}
{\sc Egan, G.}
\newblock {\em Zendegi}.
\newblock Gollancz, 2010.

\bibitem{fazelpour2020algorithmic}
{\sc Fazelpour, S., and Lipton, Z.~C.}
\newblock Algorithmic fairness from a non-ideal perspective.
\newblock In {\em Proceedings of the AAAI/ACM Conference on AI, Ethics, and
  Society\/} (2020), pp.~57--63.

\bibitem{finocchiaro2021bridging}
{\sc Finocchiaro, J., Maio, R., Monachou, F., Patro, G.~K., Raghavan, M.,
  Stoica, A.-A., and Tsirtsis, S.}
\newblock Bridging machine learning and mechanism design towards algorithmic
  fairness.
\newblock In {\em Proceedings of the 2021 ACM Conference on Fairness,
  Accountability, and Transparency\/} (2021), pp.~489--503.

\bibitem{fish2021reflexive}
{\sc Fish, B., and Stark, L.}
\newblock Reflexive design for fairness and other human values in formal
  models.
\newblock In {\em Proceedings of the 2021 AAAI/ACM Conference on AI, Ethics,
  and Society\/} (2021), pp.~89--99.

\bibitem{Fox2009VirtualST}
{\sc Fox, J., and Bailenson, J.~N.}
\newblock Virtual self-modeling: The effects of vicarious reinforcement and
  identification on exercise behaviors.
\newblock {\em Media Psychology 12\/} (2009), 1 -- 25.

\bibitem{Gatys2016}
{\sc Gatys, L.~A., Ecker, A.~S., and Bethge, M.}
\newblock {Image Style Transfer Using Convolutional Neural Networks}, 2016.

\bibitem{Geigle2017Apr}
{\sc Geigle, C., and Zhai, C.}
\newblock {Modeling MOOC Student Behavior With Two-Layer Hidden Markov Models}.
\newblock In {\em {L@S '17: Proceedings of the Fourth (2017) ACM Conference on
  Learning @ Scale}}. Association for Computing Machinery, New York, NY, USA,
  Apr 2017, pp.~205--208.

\bibitem{gitlin2021Dec}
{\sc Gitlin, J.}
\newblock {War Stories: How Forza learned to love neural nets to train AI
  drivers}, Dec 2021.
\newblock [Online; accessed 7. Dec. 2021].

\bibitem{greenberg2018william}
{\sc Greenberg, D.}
\newblock William goldman: {The} writer who brought watergate to the screen.
\newblock {\em Politico\/} (December 2018).

\bibitem{guan2018said}
{\sc Guan, M., Gulshan, V., Dai, A., and Hinton, G.}
\newblock Who said what: Modeling individual labelers improves classification.
\newblock In {\em Proceedings of the AAAI Conference on Artificial
  Intelligence\/} (2018), vol.~32.

\bibitem{guo2021detecting}
{\sc Guo, W., and Caliskan, A.}
\newblock Detecting emergent intersectional biases: Contextualized word
  embeddings contain a distribution of human-like biases.
\newblock In {\em Proceedings of the 2021 AAAI/ACM Conference on AI, Ethics,
  and Society\/} (2021), pp.~122--133.

\bibitem{guzdial2019friend}
{\sc Guzdial, M., Liao, N., Chen, J., Chen, S.-Y., Shah, S., Shah, V., Reno,
  J., Smith, G., and Riedl, M.~O.}
\newblock Friend, collaborator, student, manager: How design of an ai-driven
  game level editor affects creators.
\newblock In {\em Proceedings of the 2019 CHI conference on human factors in
  computing systems\/} (2019), pp.~1--13.

\bibitem{harcourt2015risk}
{\sc Harcourt, B.~E.}
\newblock Risk as a proxy for race: The dangers of risk assessment.
\newblock {\em Federal Sentencing Reporter 27}, 4 (2015), 237--243.

\bibitem{Hill2013AvatarEB}
{\sc Hill, D.~W.}
\newblock Avatar ethics: Beyond images and signs.
\newblock {\em Journal for Cultural Research 17\/} (2013), 69 -- 84.

\bibitem{hooshyar2018data}
{\sc Hooshyar, D., Yousefi, M., and Lim, H.}
\newblock Data-driven approaches to game player modeling: a systematic
  literature review.
\newblock {\em ACM Computing Surveys (CSUR) 50}, 6 (2018), 1--19.

\bibitem{hu2021off}
{\sc Hu, H., Lerer, A., Cui, B., Pineda, L., Brown, N., and Foerster, J.}
\newblock Off-belief learning.
\newblock In {\em International Conference on Machine Learning\/} (2021), PMLR,
  pp.~4369--4379.

\bibitem{jacob2021modeling}
{\sc Jacob, A.~P., Wu, D.~J., Farina, G., Lerer, A., Bakhtin, A., Andreas, J.,
  and Brown, N.}
\newblock Modeling strong and human-like gameplay with kl-regularized search.
\newblock {\em arXiv preprint arXiv:2112.07544\/} (2021).

\bibitem{jaderberg2019human}
{\sc Jaderberg, M., Czarnecki, W.~M., Dunning, I., Marris, L., Lever, G.,
  Castaneda, A.~G., Beattie, C., Rabinowitz, N.~C., Morcos, A.~S., Ruderman,
  A., et~al.}
\newblock Human-level performance in 3d multiplayer games with population-based
  reinforcement learning.
\newblock {\em Science 364}, 6443 (2019), 859--865.

\bibitem{jahanshahi2021auto}
{\sc Jahanshahi, H., Kazmi, S., and Cevik, M.}
\newblock Auto response generation in online medical chat services.
\newblock {\em arXiv preprint arXiv:2104.12755\/} (2021).

\bibitem{jenner2016tviv}
{\sc Jenner, M.}
\newblock Is this tviv? on netflix, tviii and binge-watching.
\newblock {\em New media \& society 18}, 2 (2016), 257--273.

\bibitem{Jia2021ScalingUV}
{\sc Jia, C., Yang, Y., Xia, Y., Chen, Y.-T., Parekh, Z., Pham, H., Le, Q.~V.,
  Sung, Y.-H., Li, Z., and Duerig, T.}
\newblock Scaling up visual and vision-language representation learning with
  noisy text supervision.
\newblock In {\em ICML\/} (2021).

\bibitem{kang2018self}
{\sc Kang, W.-C., and McAuley, J.}
\newblock Self-attentive sequential recommendation.
\newblock In {\em 2018 IEEE International Conference on Data Mining (ICDM)\/}
  (2018), IEEE, pp.~197--206.

\bibitem{kleinberg2018human}
{\sc Kleinberg, J., Lakkaraju, H., Leskovec, J., Ludwig, J., and Mullainathan,
  S.}
\newblock Human decisions and machine predictions.
\newblock {\em The quarterly journal of economics 133}, 1 (2018), 237--293.

\bibitem{kleinberg2018discrimination}
{\sc Kleinberg, J., Ludwig, J., Mullainathan, S., and Sunstein, C.~R.}
\newblock Discrimination in the age of algorithms.
\newblock {\em Journal of Legal Analysis 10\/} (2018), 113--174.

\bibitem{kober2013reinforcement}
{\sc Kober, J., Bagnell, J.~A., and Peters, J.}
\newblock Reinforcement learning in robotics: A survey.
\newblock {\em The International Journal of Robotics Research 32}, 11 (2013),
  1238--1274.

\bibitem{Koch2019MayAD}
{\sc Koch, J., Lucero, A., Hegemann, L., and Oulasvirta, A.}
\newblock May ai?: Design ideation with cooperative contextual bandits.
\newblock {\em Proceedings of the 2019 CHI Conference on Human Factors in
  Computing Systems\/} (2019).

\bibitem{kokkinakis2021metagaming}
{\sc Kokkinakis, A., York, P., Patra, M., Robertson, J., Kirman, B., Coates,
  A., Pedrassoli~Chitayat, A., Demediuk, S.~P., Drachen, A., Hook, J.~D.,
  et~al.}
\newblock Metagaming and metagames in esports.
\newblock {\em International Journal of Esports\/} (2021).

\bibitem{Korshunov2022}
{\sc Korshunov, P., and Marcel, S.}
\newblock {\em The Threat of Deepfakes to Computer and Human Visions}.
\newblock Springer International Publishing, Cham, 2022, pp.~97--115.

\bibitem{Krishnan2021Jun}
{\sc Krishnan, P., Kovvuri, R., Pang, G., Vassilev, B., and Hassner, T.}
\newblock {TextStyleBrush: Transfer of Text Aesthetics from a Single Example}.
\newblock {\em arXiv\/} (Jun 2021).

\bibitem{lazer2018science}
{\sc Lazer, D.~M., Baum, M.~A., Benkler, Y., Berinsky, A.~J., Greenhill, K.~M.,
  Menczer, F., Metzger, M.~J., Nyhan, B., Pennycook, G., Rothschild, D.,
  et~al.}
\newblock The science of fake news.
\newblock {\em Science 359}, 6380 (2018), 1094--1096.

\bibitem{Li2017Jan}
{\sc Li, Y., Wang, N., Liu, J., and Hou, X.}
\newblock {Demystifying Neural Style Transfer}.
\newblock {\em arXiv\/} (Jan 2017).

\bibitem{liang2019implicit}
{\sc Liang, C., Proft, J., Andersen, E., and Knepper, R.~A.}
\newblock Implicit communication of actionable information in human-ai teams.
\newblock In {\em Proceedings of the 2019 CHI Conference on Human Factors in
  Computing Systems\/} (2019), pp.~1--13.

\bibitem{Liu2022NameYS}
{\sc Liu, Z.-S., Wang, L.-W., Siu, W.~C., and Kalogeiton, V.~S.}
\newblock Name your style: An arbitrary artist-aware image style transfer.
\newblock {\em ArXiv abs/2202.13562\/} (2022).

\bibitem{ma2019neural}
{\sc Ma, C., Ji, Z., and Gao, M.}
\newblock Neural style transfer improves 3d cardiovascular mr image
  segmentation on inconsistent data.
\newblock In {\em International Conference on Medical Image Computing and
  Computer-Assisted Intervention\/} (2019), Springer, pp.~128--136.

\bibitem{mcauley2022personalized}
{\sc McAuley, J.}
\newblock {\em Personalized Machine Learning}.
\newblock Cambridge University Press, 2022.

\bibitem{mccarthy_2003}
{\sc McCarthy, W.}
\newblock {\em The wellstone}.
\newblock Bantam Books, 2003.

\bibitem{mcilroy2020aligning}
{\sc McIlroy-Young, R., Sen, S., Kleinberg, J., and Anderson, A.}
\newblock Aligning superhuman ai with human behavior: Chess as a model system.
\newblock In {\em Proceedings of the 26th ACM SIGKDD International Conference
  on Knowledge Discovery \& Data Mining\/} (2020), pp.~1677--1687.

\bibitem{mcilroy2020learning}
{\sc McIlroy-Young, R., Wang, R., Sen, S., Kleinberg, J., and Anderson, A.}
\newblock Learning models of individual human behavior in chess.
\newblock In {\em Proceedings of the 28th ACM SIGKDD international conference
  on Knowledge discovery and data mining\/} (2022).

\bibitem{mehrabi2021survey}
{\sc Mehrabi, N., Morstatter, F., Saxena, N., Lerman, K., and Galstyan, A.}
\newblock A survey on bias and fairness in machine learning.
\newblock {\em ACM Computing Surveys (CSUR) 54}, 6 (2021), 1--35.

\bibitem{Melhart}
{\sc Melhart, D., Azadvar, A., Canossa, A., Liapis, A., and Yannakakis, G.~N.}
\newblock {Your Gameplay Says It All: Modelling Motivation in Tom Clancy{'}s
  The Division}.
\newblock In {\em {2019 IEEE Conference on Games (CoG)}}. IEEE, 2019,
  pp.~20--23.

\bibitem{mitchell2019model}
{\sc Mitchell, M., Wu, S., Zaldivar, A., Barnes, P., Vasserman, L., Hutchinson,
  B., Spitzer, E., Raji, I.~D., and Gebru, T.}
\newblock Model cards for model reporting.
\newblock In {\em Proceedings of the conference on fairness, accountability,
  and transparency\/} (2019), pp.~220--229.

\bibitem{mnih2015human}
{\sc Mnih, V., Kavukcuoglu, K., Silver, D., Rusu, A.~A., Veness, J., Bellemare,
  M.~G., Graves, A., Riedmiller, M., Fidjeland, A.~K., Ostrovski, G., et~al.}
\newblock Human-level control through deep reinforcement learning.
\newblock {\em nature 518}, 7540 (2015), 529--533.

\bibitem{moravvcik2017deepstack}
{\sc Morav{\v{c}}{\'\i}k, M., Schmid, M., Burch, N., Lis{\`y}, V., Morrill, D.,
  Bard, N., Davis, T., Waugh, K., Johanson, M., and Bowling, M.}
\newblock Deepstack: Expert-level artificial intelligence in heads-up no-limit
  poker.
\newblock {\em Science 356}, 6337 (2017), 508--513.

\bibitem{morgenstern1953theory}
{\sc Morgenstern, O., and Von~Neumann, J.}
\newblock {\em Theory of games and economic behavior}.
\newblock Princeton university press, 1953.

\bibitem{Morris2013Aug}
{\sc Morris, C.}
\newblock {Former NCAA athletes win video game lawsuit against EA}.
\newblock {\em NBC News\/} (Aug 2013).

\bibitem{narayanan2008robust}
{\sc Narayanan, A., and Shmatikov, V.}
\newblock Robust de-anonymization of large sparse datasets.
\newblock In {\em 2008 IEEE Symposium on Security and Privacy (sp 2008)\/}
  (2008), IEEE, pp.~111--125.

\bibitem{ng2001discriminative}
{\sc Ng, A., and Jordan, M.}
\newblock On discriminative vs. generative classifiers: A comparison of
  logistic regression and naive bayes.
\newblock {\em Advances in neural information processing systems 14\/} (2001).

\bibitem{park2019gaugan}
{\sc Park, T., Liu, M.-Y., Wang, T.-C., and Zhu, J.-Y.}
\newblock Gaugan: Semantic image synthesis with spatially adaptive
  normalization.
\newblock In {\em ACM SIGGRAPH 2019 Real-Time Live!\/} (New York, NY, USA,
  2019), SIGGRAPH '19, Association for Computing Machinery.

\bibitem{Peck2013PuttingYI}
{\sc Peck, T.~C., Seinfeld, S., Aglioti, S.~M., and Slater, M.}
\newblock Putting yourself in the skin of a black avatar reduces implicit
  racial bias.
\newblock {\em Consciousness and Cognition 22\/} (2013), 779--787.

\bibitem{pohl1977gateway}
{\sc Pohl, F.}
\newblock {\em Gateway}.
\newblock St. Martin's Press, 1977.

\bibitem{dalletweet}
{\sc Porres, D.}
\newblock A mayan warrior getting ready, in the style of rembrandt.
\newblock \url{https://twitter.com/PDillis/status/1530297800453496833}, May
  2022.

\bibitem{Qi2020ImageBERTCP}
{\sc Qi, D., Su, L., Song, J., Cui, E., Bharti, T., and Sacheti, A.}
\newblock Imagebert: Cross-modal pre-training with large-scale weak-supervised
  image-text data.
\newblock {\em ArXiv abs/2001.07966\/} (2020).

\bibitem{radford2021learning}
{\sc Radford, A., Kim, J.~W., Hallacy, C., Ramesh, A., Goh, G., Agarwal, S.,
  Sastry, G., Askell, A., Mishkin, P., Clark, J., et~al.}
\newblock Learning transferable visual models from natural language
  supervision.
\newblock In {\em International Conference on Machine Learning\/} (2021), PMLR,
  pp.~8748--8763.

\bibitem{raghavan2020mitigating}
{\sc Raghavan, M., Barocas, S., Kleinberg, J., and Levy, K.}
\newblock Mitigating bias in algorithmic hiring: Evaluating claims and
  practices.
\newblock In {\em Proceedings of the 2020 conference on fairness,
  accountability, and transparency\/} (2020), pp.~469--481.

\bibitem{ramazzotti2018encoding}
{\sc Ramazzotti, M., Buscema, P.~M., Massini, G., and Della~Torre, F.}
\newblock Encoding and simulating the past. a machine learning approach to the
  archaeological information.
\newblock In {\em 2018 Metrology for Archaeology and Cultural Heritage
  (MetroArchaeo)\/} (2018), IEEE, pp.~39--44.

\bibitem{ramesh2022hierarchical}
{\sc Ramesh, A., Dhariwal, P., Nichol, A., Chu, C., and Chen, M.}
\newblock Hierarchical text-conditional image generation with clip latents.
\newblock {\em arXiv preprint arXiv:2204.06125\/} (2022).

\bibitem{Ramesh2021ZeroShotTG}
{\sc Ramesh, A., Pavlov, M., Goh, G., Gray, S., Voss, C., Radford, A., Chen,
  M., and Sutskever, I.}
\newblock Zero-shot text-to-image generation.
\newblock {\em ArXiv abs/2102.12092\/} (2021).

\bibitem{ranzini2017love}
{\sc Ranzini, G., and Lutz, C.}
\newblock Love at first swipe? explaining tinder self-presentation and motives.
\newblock {\em Mobile Media \& Communication 5}, 1 (2017), 80--101.

\bibitem{resnick1997recommender}
{\sc Resnick, P., and Varian, H.~R.}
\newblock Recommender systems.
\newblock {\em Communications of the ACM 40}, 3 (1997), 56--58.

\bibitem{ressmeyer2019deep}
{\sc Ressmeyer, R., Masling, S., and Liao, M.}
\newblock “deep faking” political twitter using transfe r learning and
  gpt-2, 2019.

\bibitem{rezaee2008corporate}
{\sc Rezaee, Z.}
\newblock {\em Corporate governance and ethics}.
\newblock John Wiley \& Sons, 2008.

\bibitem{Rosner2021Jul}
{\sc Rosner, H.}
\newblock {The Ethics of a Deepfake Anthony Bourdain Voice in
  {\textquotedblleft}Roadrunner{\textquotedblright}}.
\newblock {\em New Yorker\/} (Jul 2021).

\bibitem{sanakoyeu2018style}
{\sc Sanakoyeu, A., Kotovenko, D., Lang, S., and Ommer, B.}
\newblock A style-aware content loss for real-time hd style transfer.
\newblock In {\em proceedings of the European conference on computer vision
  (ECCV)\/} (2018), pp.~698--714.

\bibitem{selbst2019fairness}
{\sc Selbst, A.~D., Boyd, D., Friedler, S.~A., Venkatasubramanian, S., and
  Vertesi, J.}
\newblock Fairness and abstraction in sociotechnical systems.
\newblock In {\em Proceedings of the conference on fairness, accountability,
  and transparency\/} (2019), pp.~59--68.

\bibitem{silver2016mastering}
{\sc Silver, D., Huang, A., Maddison, C.~J., Guez, A., Sifre, L., Van
  Den~Driessche, G., Schrittwieser, J., Antonoglou, I., Panneershelvam, V.,
  Lanctot, M., et~al.}
\newblock Mastering the game of go with deep neural networks and tree search.
\newblock {\em nature 529}, 7587 (2016), 484--489.

\bibitem{silver2018general}
{\sc Silver, D., Hubert, T., Schrittwieser, J., Antonoglou, I., Lai, M., Guez,
  A., Lanctot, M., Sifre, L., Kumaran, D., Graepel, T., et~al.}
\newblock A general reinforcement learning algorithm that masters chess, shogi,
  and go through self-play.
\newblock {\em Science 362}, 6419 (2018), 1140--1144.

\bibitem{simmons2020catfishing}
{\sc Simmons, M., and Lee, J.~S.}
\newblock Catfishing: A look into online dating and impersonation.
\newblock In {\em International Conference on Human-Computer Interaction\/}
  (2020), Springer, pp.~349--358.

\bibitem{mtgMetaSolved}
{\sc Stein, R.}
\newblock {What We Learned{\ifmmode---\else\textemdash\fi}Solving Standard -
  Hipsters of the Coast}, Nov 2015.

\bibitem{Suh2011WhatIY}
{\sc Suh, K.-S., Kim, H., and Suh, E.-K.}
\newblock What if your avatar looks like you? dual-congruity perspectives for
  avatar use.
\newblock {\em MIS Q. 35\/} (2011), 711--729.

\bibitem{taddeo2018ai}
{\sc Taddeo, M., and Floridi, L.}
\newblock How ai can be a force for good.
\newblock {\em Science 361}, 6404 (2018), 751--752.

\bibitem{Tantaros2019Jan}
{\sc Tantaros, A.}
\newblock {Electronic Arts, identity thief?}
\newblock {\em Nydailynews\/} (Jan 2019).

\bibitem{Thomas2010Nov}
{\sc Thomas, K.}
\newblock {Sports Video Game Suit Gets to Heart of First Amendment Clash}.
\newblock {\em N.Y. Times\/} (Nov 2010).

\bibitem{TOLOSANA2020131}
{\sc Tolosana, R., Vera-Rodriguez, R., Fierrez, J., Morales, A., and
  Ortega-Garcia, J.}
\newblock Deepfakes and beyond: A survey of face manipulation and fake
  detection.
\newblock {\em Information Fusion 64\/} (2020), 131--148.

\bibitem{tomavsev2020assessing}
{\sc Toma{\v{s}}ev, N., Paquet, U., Hassabis, D., and Kramnik, V.}
\newblock Assessing game balance with alphazero: Exploring alternative rule
  sets in chess.
\newblock {\em arXiv preprint arXiv:2009.04374\/} (2020).

\bibitem{deepfake}
{\sc Tora, M.}
\newblock Faceswap, 2018.

\bibitem{vaccari2020deepfakes}
{\sc Vaccari, C., and Chadwick, A.}
\newblock Deepfakes and disinformation: Exploring the impact of synthetic
  political video on deception, uncertainty, and trust in news.
\newblock {\em Social Media+ Society 6}, 1 (2020), 2056305120903408.

\bibitem{vinge_monster}
{\sc Vinge, V.}
\newblock {\em The Cookie Monster}.
\newblock Analog Science Fiction and Fact, 2003.

\bibitem{whittaker2019disability}
{\sc Whittaker, M., Alper, M., Bennett, C.~L., Hendren, S., Kaziunas, L.,
  Mills, M., Morris, M.~R., Rankin, J., Rogers, E., Salas, M., et~al.}
\newblock Disability, bias, and ai.
\newblock {\em AI Now Institute\/} (2019).

\bibitem{Wolfendale2006MyAM}
{\sc Wolfendale, J.}
\newblock My avatar, my self: Virtual harm and attachment.
\newblock {\em Ethics and Information Technology 9\/} (2006), 111--119.

\bibitem{wu2019session}
{\sc Wu, S., Tang, Y., Zhu, Y., Wang, L., Xie, X., and Tan, T.}
\newblock Session-based recommendation with graph neural networks.
\newblock In {\em Proceedings of the AAAI conference on artificial
  intelligence\/} (2019), vol.~33, pp.~346--353.

\bibitem{xu2022deep}
{\sc Xu, P., Hospedales, T.~M., Yin, Q., Song, Y.-Z., Xiang, T., and Wang, L.}
\newblock Deep learning for free-hand sketch: A survey.
\newblock {\em IEEE Transactions on Pattern Analysis and Machine
  Intelligence\/} (2022).

\bibitem{Yee2007ThePE}
{\sc Yee, N., and Bailenson, J.~N.}
\newblock The proteus effect: The effect of transformed self-representation on
  behavior.
\newblock {\em Human Communication Research 33\/} (2007), 271--290.

\bibitem{Yee2009ImplicationsOT}
{\sc Yee, N., Bailenson, J.~N., and Ducheneaut, N.}
\newblock Implications of transformed digital self-representation on online and
  offline behavior.
\newblock {\em Communication Research 36\/} (2009), 285--312.

\bibitem{Yee2009ThePE}
{\sc Yee, N., Bailenson, J.~N., and Ducheneaut, N.}
\newblock The proteus effect.
\newblock {\em Communication Research 36\/} (2009), 285 -- 312.

\bibitem{zellers2019defending}
{\sc Zellers, R., Holtzman, A., Rashkin, H., Bisk, Y., Farhadi, A., Roesner,
  F., and Choi, Y.}
\newblock Defending against neural fake news.
\newblock {\em Advances in neural information processing systems 32\/} (2019).

\bibitem{Zhang2018Jan}
{\sc Zhang, S., Dinan, E., Urbanek, J., Szlam, A., Kiela, D., and Weston, J.}
\newblock {Personalizing Dialogue Agents: I have a dog, do you have pets too?}
\newblock {\em arXiv\/} (Jan 2018).

\bibitem{zimmer2020but}
{\sc Zimmer, M.}
\newblock “but the data is already public”: on the ethics of research in
  facebook.
\newblock In {\em The Ethics of Information Technologies}. Routledge, 2020,
  pp.~229--241.

\bibitem{zwitter2014big}
{\sc Zwitter, A.}
\newblock Big data ethics.
\newblock {\em Big Data \& Society 1}, 2 (2014), 2053951714559253.

\end{thebibliography}
